\documentclass[10pt,twocolumn,letterpaper]{article}

\usepackage[pagenumbers]{cvpr} 
\usepackage{multirow}
\definecolor{cblue}{rgb}{0.21,0.49,0.74}
\usepackage[pagebackref,breaklinks,colorlinks,allcolors=cblue]{hyperref}

\title{SoftShadow: Leveraging Soft Masks for Penumbra-Aware Shadow Removal}

\author {
    Xinrui Wang\textsuperscript{1}\thanks{Equal contribution for both authors} \quad
    Lanqing Guo\textsuperscript{2}\footnotemark[1] \quad
    Xiyu Wang\textsuperscript{1} \quad
    Siyu Huang\textsuperscript{3} \quad
    Bihan Wen\textsuperscript{1}\\
    \textsuperscript{1}Nanyang Technological University, Singapore\\
    \textsuperscript{2}The University of Texas at Austin, USA\\
    \textsuperscript{3}Clemson University, USA
}

\begin{document}
\maketitle

\begin{abstract}
Recent advancements in deep learning have yielded promising results for the image shadow removal task.
However, most existing methods rely on binary pre-generated shadow masks. The binary nature of such masks could potentially lead to artifacts near the boundary between shadow and non-shadow areas. 
In view of this, inspired by the physical model of shadow formation, we introduce novel soft shadow masks specifically designed for shadow removal. 
To achieve such soft masks, we propose a SoftShadow framework by leveraging the prior knowledge of pretrained SAM and integrating physical constraints. Specifically, we jointly tune the SAM and the subsequent shadow removal network using penumbra formation constraint loss, mask reconstruction loss, and shadow removal loss. 
This framework enables accurate predictions of penumbra (partially shaded) and umbra (fully shaded) areas while simultaneously facilitating end-to-end shadow removal. Through extensive experiments on popular datasets, we found that our SoftShadow framework, which generates soft masks, can better restore boundary artifacts, achieve state-of-the-art performance, and demonstrate superior generalizability.

\end{abstract}

\section{Introduction}{\label{intro}}


Shadow removal aims to restore content obscured in shadow regions and correct degraded illumination. Recently, deep learning methods have shown excellent performance in shadow removal tasks relying on large-scale training data. However, one of the main challenges in shadow removal arises in images with soft shadows, where the shadow edges are blurred rather than sharp. This blurred area, known as the \textit{penumbra area}, occurs due to partial occlusion of light, creating a transition zone where illumination varies significantly. Removing shadows in these regions is challenging because the gradual light-to-shadow transition often leads to boundary artifacts, compromising the quality of the shadow removal results. 



\begin{figure}[t!]
    \centering
    \includegraphics[width=0.47\textwidth]{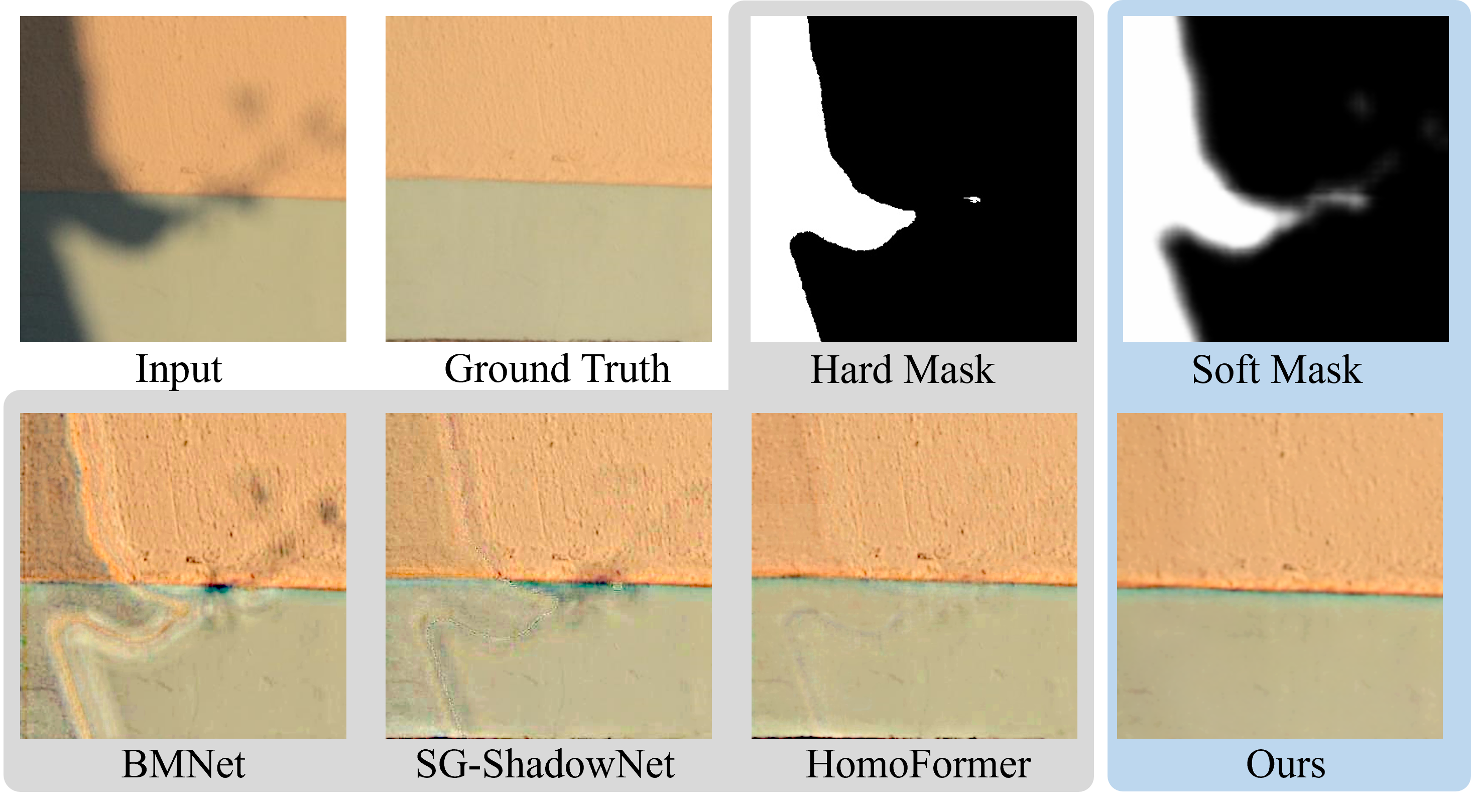}
    \caption{Illustration of soft shadow removal results using our proposed \textit{SoftShadow} with \textit{soft} shadow mask compared to the recent competing methods BMNet~\cite{zhu2022bijective}, SG-ShadowNet~\cite{wan2022style} and HomoFormer~\cite{xiao2024homoformer} using \textit{hard} shadow masks. The second rows are sharpened versions of results for better visualization. }
    \label{intro}
\end{figure}

A few methods rely on hard shadow masks (binary masks) to indicate the shadow regions~\cite{zhu2022bijective, guo2023shadowformer, guo2023shadowdiffusion, xiao2024homoformer}.  These binary shadow masks are either manually annotated~\cite{wang2018stacked, liu2024recasting}, or predicted by off-the-shelf shadow detectors~\cite{cun2020towards}. Obtaining such shadow masks can be costly and complicated, and the choice of shadow detector can significantly impact shadow removal performance. Moreover, hard masks fall short of representing the penumbra area, causing boundary artifacts in removal results, especially when the penumbra areas are highly pronounced.  As shown in Figure~\ref{intro}, when the input image contains penumbra regions, the previous state-of-the-art (SOTA) method~\cite{xiao2024homoformer} suffers from boundary artifacts. 
Other approaches attempt shadow removal without explicitly extracting the shadow mask. Instead, they incorporate semantic parsing modules into shadow removal networks. For instance, some approaches involve predicting degradation attention~\cite{qu2017deshadownet,cun2020towards}, while others utilize domain classifiers~\cite{jin2021dc,jin2024des3} to better understand and address shadow effects. However, their performance may be limited by the lack of powerful external detectors providing additional shadow location information.

In view of this, we argue that simply leveraging the binary mask to represent the shadow location is not enough. Differently, we introduce a novel soft shadow mask (grayscale mask) specifically designed for shadow removal as shown in the hard mask \textit{v.s.} soft mask in Figure~\ref{intro}. 
The soft masks can precisely locate the inner and outer boundaries of shadow and indicate the penumbra area with proper degradation variance. 
Moreover, recognizing the effectiveness of employing a powerful pretrained detector for providing semantic information, we utilize the prior knowledge from the pretrained SAM~\cite{kirillov2023segment}. While existing methods~\cite{zhang2023sam} have used the off-the-shelf SAM in a naive way, merely predicting the shadow mask as input to the shadow removal network, we adapt SAM as a soft shadow mask predictor and jointly optimize it with the shadow removal network.

\begin{figure}[t!]
    \centering
    \includegraphics[width=0.47\textwidth]{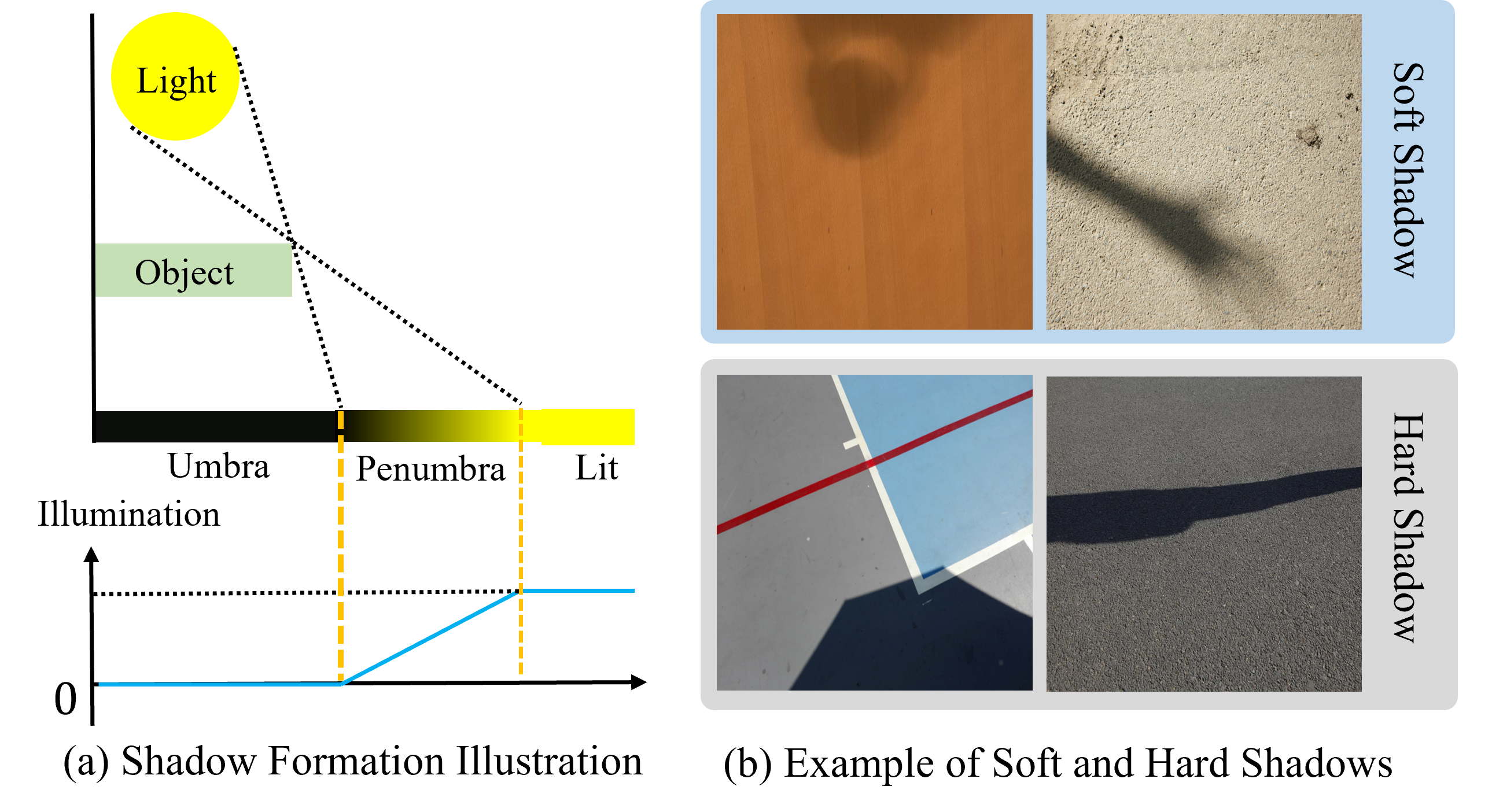}
    \caption{(a) Illustration of the shadow formation geometry that creates the penumbra and umbra~\cite{nielsen2007segmentation} regions; the umbra area is where the light is fully occluded, the penumbra area is where light is partially occluded, the lit area is where light is not occluded. (b) Examples of soft shadow images and hard shadow images from commonly used datasets.}
    \label{soft hard examples}
\end{figure}
In this paper, we first propose a unified shadow removal framework, dubbed \textit{SoftShadow}, which removes shadow from images in an end-to-end manner without requiring input shadow masks as shown in Figure~\ref{main framework}.
We integrate the pretrained SAM~\cite{kirillov2023segment} combined with Low-Rank Adaptation (LoRA)~\cite{hu2021lora} to jointly adapt the SAM model and shadow removal model. 
The framework enables accurate predictions of soft shadow masks as intermediate results while simultaneously facilitating end-to-end shadow removal.
Besides, we introduce a penumbra formation constraint to assist predict soft shadow masks. This constraint regularizes the gradient of the predicted mask in the penumbra area, resulting in more accurate predictions with detailed and spatially varied shadow masks.
In the commonly used shadow removal datasets, SRD and LRSS datasets contain soft shadow images, which makes them more suitable for illustrating the improvements of utilizing soft shadow masks. Experimental results show that \textit{SoftShadow} achieves superior performance on SRD and LRSS datasets. Our main contributions are summarized as follows:

\begin{itemize}
    \item We propose a unified shadow removal framework that does not require additional input shadow masks. This framework enables accurate predictions of soft masks as intermediate results, which is specifically designed to capture detailed and spatially varied shadow location information.
    \item We introduce penumbra formation constraint inspired by the physical shadow formation model to further refine the soft mask in the penumbra area. By leveraging the constraint loss and shadow removal loss, we jointly tune the SAM and the subsequent shadow removal network.
        
    
    \item Experimental results demonstrate that SoftShadow surpasses state-of-the-art shadow removal methods on the SRD and LRSS datasets, achieving superior performance and even comparable results with previous methods that use ground truth mask inputs.
\end{itemize}

\section{Related Work}

\subsection{Shadow Removal}
The degradation of shadows varies in each image, posing a significant challenge for shadow image restoration. In recent years deep learning-based approaches have achieved remarkable results in shadow removal. 
Some methods restore the shadow image with the guidance of shadow masks. For example, SP+M-Net~\cite{le2019shadow} employs two deep networks to predict shadow matte and shadow parameters. Recently, powerful backbones such as transformers~\cite{vaswani2017attention, dosovitskiy2020image}, and diffusion models \cite{ho2020denoising, saharia2022image} have been applied to the shadow removal task. HomoFormer~\cite{xiao2024homoformer} homogenizes the spatial distribution of shadow masks to uniformly recover the entire shadow image, while ShadowDiffusion~\cite{guo2023shadowdiffusion} provides a robust generative prior for producing natural shadow-free images. 
Some methods aim to eliminate the dependency on shadow mask inputs. DeShadowNet~\cite{qu2017deshadownet} introduces an end-to-end shadow removal method containing a multi-branch fusion module. ST-CGAN~\cite{wang2018stacked} connects two GANs~\cite{goodfellow2014generative} in sequence to jointly detect and remove shadows. DC-ShadowNet~\cite{jin2021dc} provides an unsupervised domain-classifier discriminator for guided shadow removal network. More recently, DeS3~\cite{jin2024des3} implemented a method capable of removing shadows that are cast on the object itself without requiring shadow mask inputs. However, due to the absence of additional shadow location information, their performance may degrade, and the number of parameters used could increase significantly. 
Different from these existing methods, we introduce a new concept ``soft mask'',  designed explicitly for shadow removal.

\begin{figure*}[t]
    \centering
    \includegraphics[width=0.80\textwidth]{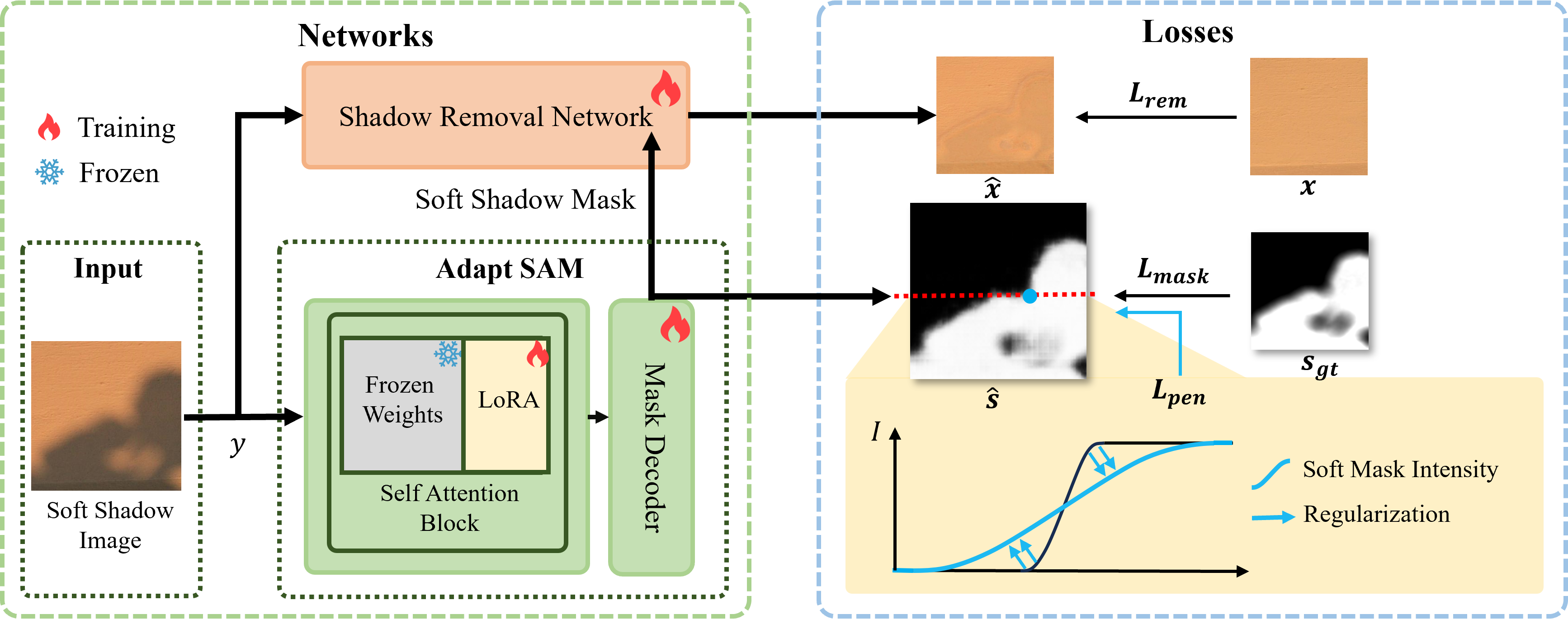}
    \caption{Illustration of the proposed SoftShadow. The \textbf{left box} illustrates the SoftShadow networks, where a shadow image $\mathbf{y}$ is input into SAM for detecting soft shadow masks. The shadow removal network then processes the soft mask and the shadow image to produce a shadow-free image. The \textbf{right box} shows the three losses we used in SoftShadow. From top to bottom, the shadow removal loss $\mathcal{L}_{rem}$ is calculated between shadow-free images and shadow-removal images. The mask reconstruction loss $\mathcal{L}_{mask}$ is calculated between predicted soft masks and ours ground truth soft masks. The penumbra formation constraint loss $\mathcal{L}_{pen}$ act as a regularization term. It aims to regularize the gradient of predicted soft masks in the penumbra area. As shown in the soft mask intensity curve, the ideal mask intensity in the penumbra area should not be too large and the gradient orientation should be consistence. }
    \label{main framework}
\end{figure*}


\subsection{Shadow Detection}
Shadow detection can be approached as a segmentation task, it often struggles with small shadows and indistinct shadow edges. Various methods have been proposed to address these challenges. BDRAR~\cite{zhu2018bidirectional} introduces a bidirectional pyramidal architecture for shadow detection. 
DSD~\cite{zheng2019distraction} designs a distraction-aware module to minimize false positives in shadow detection. 
Chen et al. \cite{chen2020multi} propose a task-specific semi-supervised learning mechanism to utilize unlabeled shadow images for detection, enhancing the robustness of the model. DHAN~\cite{cun2020towards} is a well-used shadow detection model using dual hierarchical aggregation network training on synthetic shadow and shadow-free image pairs to improve detection accuracy. The SAM-Adapter~\cite{chen2023sam} has been utilized to fine-tune the Segment Anything Model (SAM)~\cite{kirillov2023segment} specifically for shadow detection tasks. While this approach improves accuracy, it demands a substantial number of model parameters. However, most existing methods detect shadows as hard masks and fail to represent the penumbra area. In contrast, our approach leverages the shadow removal loss and the Penumbra Formation Constraint loss, jointly training the shadow removal network and the shadow detection network.

\section{Methods}

In this section, we first explain the motivation behind our newly introduced soft shadow mask, inspired by the physical model of  shadow formation. We then propose the unified shadow removal framework \textit{SoftShadow}, which leverages the powerful segmentation capabilities of SAM to produce the soft shadow mask and guide the removal network to generate better shadow-free images. Finally, to further mitigate boundary artifacts, we introduce the penumbra formation constraint loss which ensures a smoother transition between the shadow region and the non-shadow region, offering better guidance to the shadow removal network.


\subsection{Motivation}



Shadow degradations in the real world exhibit considerable amount of variation and can be classified into two main categories: soft shadows and hard shadows~\cite{langer1997light}. This classification depends on the light source and the distance between the object and the surface, as shown in Figure~\ref{soft hard examples}(a). Soft shadows are characterized by blurry edges and gradual transitions from light to dark, creating relatively large penumbra areas, while hard shadows have sharp edges with few to none penumbra regions.

Previous methods~\cite{zhu2022bijective,fu2021auto,guo2023shadowformer,xiao2024homoformer} commonly use hard masks to guide the shadow removal process. The hard masks effectively provide positional guidance for hard masks with sharp edges. However, when the penumbra regions are more pronounced, as shown in Figure~\ref{soft hard examples} (b),  the boundary between the umbra and the lit region is hard to determine, making hard masks unsuitable for representing such soft shadows.
Additionally, we observed that, in general, the brightness transition within the penumbra area can vary significantly, whereas the brightness of the lit and the umbra areas are relatively uniform.

To this end, we instead use soft (greyscale) masks to represent the position of the shadows. Formally, we denote the soft mask as $\mathbf{s}$. For a given shadow-free image $\mathbf{x}$, the shadow image $\mathbf{y}$ can be formulated as such:
\begin{equation}
    \mathbf{y} = \mathbf{a}\cdot \mathbf{s}\cdot \mathbf{x} + (\mathbf{1}-\mathbf{s})\cdot \mathbf{x},
\end{equation}
where $\cdot $ denotes element-wise multiplication, $\mathbf{s}$ is the soft shadow mask in which constant $s=0$ represents lit area, $s=1$ represents umbra area, and $s\in[0,1]$ represents penumbra area. $a \in [0,1]$ represents the illumination weight in the non-lit area. Based on our experiment, we quantitatively and qualitatively found that the proposed shadow degradation model could lead to fewer boundary artifacts and generally better shadow removal results.

\subsection{Architecture of SoftShadow }

SoftShadow is a unified framework designed to remove object shadows in images in an end-to-end manner. The overall architecture is illustrated in Figure~\ref{main framework}. SoftShadow contains a soft shadow detector and a subsequent shadow removal network. The soft shadow detector leverages the strong segmentation capabilities of SAM and further finetuning it to serve as a soft mask detector.  This is because the pretrained SAM can fail to accurately identify shadows~\cite{kirillov2023segment, jie2023sam}. Moreover, since SAM has noticeably more parameters compared to conventional shadow detection networks, we use Low-Rank Adaptation (LoRA) \cite{hu2021lora} to reduce the number of trainable parameters. In practice, we tune all the self-attention blocks of image encoders using LoRA. Since the mask decoder is lightweight, we tune all the parameters of it. 

SAM is mainly trained for binary predictions, unlike the soft masks we aim to generate. To leverage SAM to generate continuous soft masks, we further introduce the mask reconstruction loss $\mathcal{L}_{mask}$ and the penumbra formation constraint loss $\mathcal{L}_{pen}$ to model the characteristics of soft masks (details are described in Section~\ref{sec:penumbra}). They encourage the model to detect the shadow area and produce a continuous soft mask rather than a binary one, thereby capturing the shadow position more accurately. We formulate a mask reconstruction loss $\mathcal{L}_{mask}$ as follows: 
\begin{equation}
        \mathcal{L}_{mask} = \mathbb{E}_n  \left\| \hat{\mathbf{s}} - \mathbf{s}_{{gt}} \right\|_F^2\;,
\end{equation}
where $n$ is the index of shadow masks, $\hat{\mathbf{s}}$ is the predicted soft shadow mask, and $\mathbf{s}_{{gt}}$ is the ground truth soft shadow mask achieved by dividing the shadow-free image $\mathbf{y}$ by the shadow image $\mathbf{x}$. 
Specifically, we convert $\mathbf{y}$ and $\mathbf{x}$ into YCbCr image space, using the Y channel to illustrate the brightness of these images, denoted as $\mathbf{y}_{Y}$ and $\mathbf{x}_{Y}$, respectively. We then divide the $\mathbf{y}_{Y}$ by $\mathbf{x}_{Y}$ to obtain the difference between them, apply a low-pass filter $f$ to reduce noise, and use a threshold to eliminate outliers as follows:  

\begin{equation}
    \mathbf{s}_{{gt}} = \text{max}\left(t, f\left(\frac{ \mathbf{x}_{Y}}{ \mathbf{y}_{Y}}\right)\right)\;,
\end{equation}
where $t$ is the threshold used to define the outer boundary between the penumbra area and the lit area.

Moreover, as illustrate in Figure~\ref{main framework}, we incorpotare the gradient of the shadow removal loss $\mathcal{L}_{rem}$ to jointly train SAM and the subsequent shadow removal network:
\begin{equation}
    \mathcal{L}_{rem} = \mathbb{E}_n \left\| \mathbf{\hat{x}} - \mathbf{x} \right\|_F^2,
\end{equation}
where $n$ indexes all images, $\hat{\mathbf{x}}$ is the restored shadow-free image, and $\mathbf{x}$ is the ground truth shadow-free image.
To this end, the intermediate soft shadow masks can be refined using shadow removal results, with the SAM adapter optimized for better location guidance in subsequent shadow removal. 

The overall training objective of SoftShadow is:
\begin{equation}
    \mathcal{L} = \mathcal{L}_{mask}+\lambda_1\mathcal{L}_{pen}+\lambda_2\mathcal{L}_{rem},
\end{equation}
where $\lambda_1$ and $\lambda_2$ are weighting coefficients to balance the influence of each term. With the merits of joint training and physical constraints, our SoftShadow framework effectively removes shadows. 




\begin{figure}[t!]
    \centering
    \includegraphics[width=0.32\textwidth]{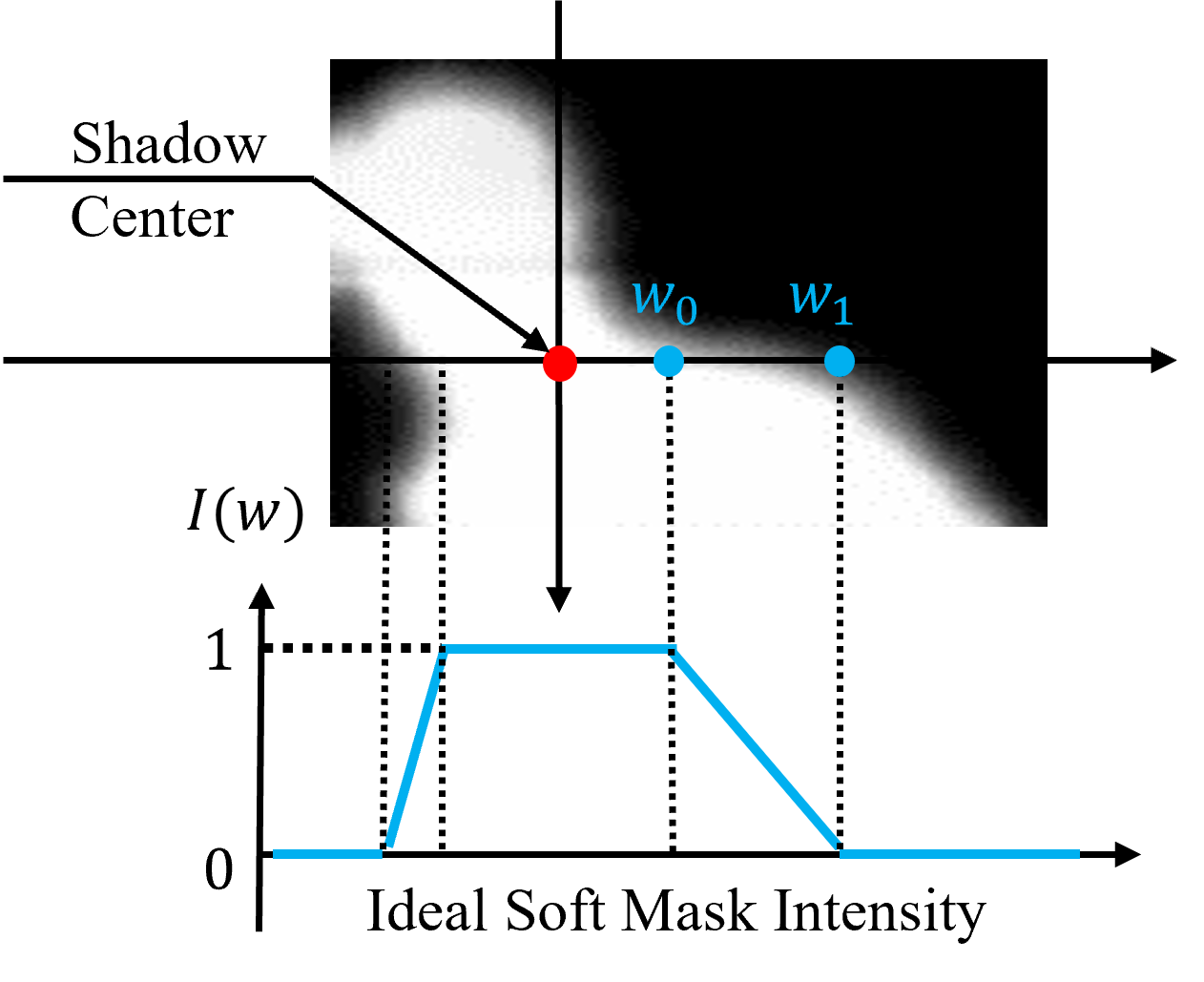}
    \caption{Illustrate the concept of Penumbra Formation Constraint. The $w_0$ and $w_1$ are points in the inner boundary and the outer boundary of the penumbra area, respectively. In the ideal soft shadow mask, the intensity in the penumbra area decreases progressively from the shadow center to the umbra area. The gradient orientation points from the shadow center towards the non-shadow area.}
    \label{penumbra constraint}
\end{figure}
 
\subsection{Penumbra Formation Constraint}\label{sec:penumbra}
We introduce the Penumbra Formation Constraint to enable SAM to predict the soft masks more accurately. The objective of this constraint is to regularize SAM to learn a mask that can reflect the illuminance change of shadow images. To meet the objective, we present two assumptions for our proposed soft mask. 
For a given shadow image, the soft mask intensity in the penumbra region should gradually transition outward from the shadow center to the outer edge of the shadow. This means that the gradient direction in the penumbra should point from the shadow center toward the boundary of the shadow. Additionally, since the intensity change in this area is gradual, the gradient magnitude should remain relatively small to reflect a smooth transition.  Based on these assumptions, we can formulate the penumbra formation constraint.
For a predicted soft mask $\mathbf{s}$, we define the penumbra area using two thresholds $t_1$ and $t_2$ as  
\begin{equation}
\mathbf{w} = \{ (i, j) \mid t_1 \leq \mathbf{s}_{i,j} \leq t_2 \},
\end{equation}
where $i$ and $j$ denote the spatial coordinates of pixels within the soft mask $\mathbf{s}$. As shown in Figure~\ref{penumbra constraint}, the center of the shadow area is denoted as $\mathbf{c}$. $\mathbf{c}$ is defined as the mean of the $x$- and $y$- coordinates of the penumbra area. The unit vector representing the direction from $\mathbf{c}$ to $\mathbf{w}$ in the penumbra area can be expressed as: 
\begin{equation}
     \mathbf{d}(\mathbf{w}) = \frac{\mathbf{w}-\mathbf{c}}{\|\mathbf{w}-\mathbf{c}\|}, \quad \text{for }\mathbf{w} \neq  \mathbf{c}
\end{equation}
Following the assumptions we proposed, we can define the penumbra formation constraint loss $\mathcal{L}_{pen}$ as:
\begin{equation}
\mathcal{L}_{\text{pen}} = \mathbb{E}_{n, \mathbf{w}} \left[ R\left(\mathbf{d}(\mathbf{w}) \cdot \nabla M(\mathbf{w}) \right) \right],
\end{equation}
where $n$ is the index of shadow masks, the desired direction of gradient is $\mathbf{d}(\mathbf{w})$, and $\cdot$ indicate the dot product. $R$ refers to the ReLU function~\cite{nair2010rectified}.
The ReLU function is applied to filter out gradients conflict with the desired direction. The constraint loss regularizes the magnitude of gradients in alignment with the desired direction, promoting a smooth intensity transition within the penumbra region. By penalizing quick changes in intensity, this constraint encourages the detector to generate softer shadow masks.

\begin{table*}[t!]
\centering
\small
\renewcommand{\arraystretch}{0.95}
\resizebox{0.99\textwidth}{!}
{%
\begin{tabular}{c|c|c|ccc|ccc|ccc}
\hline\hline
& \multirow{2}{*}{Methods} &\multirow{2}{*}{Input Masks}& \multicolumn{3}{c|}{shadow} & \multicolumn{3}{c|}{non-shadow} & \multicolumn{3}{c}{all} \\
\cline{4-12}
&  && PSNR$\uparrow$& SSIM$\uparrow$& MAE$\downarrow$& PSNR$\uparrow$& SSIM$\uparrow$& MAE$\downarrow$& PSNR$\uparrow$& SSIM$\uparrow$& MAE$\downarrow$\\
\hline
\multirow{11}{*}{\rotatebox[origin=c]{90}{SRD}} 
 & DHAN &DHAN& 33.67& 0.978&  8.94& 34.79& 0.979& 4.80& 30.51& 0.949&5.67\\
 & DC-ShadowNet &N/A& 34.00& 0.975& 7.70& 35.53& 0.981& 3.65& 31.53& 0.955&4.65\\
 & BMNet &DHAN & 35.05& 0.981& 6.61& 36.02& 0.982& 3.61& 31.69& 0.956&4.46\\
 & SG-ShadowNet &DHAN & 33.73& 0.979& 7.53& 36.18& 0.982& 2.97& 31.16& 0.952&4.23\\
& ShadowFormer &DHAN & 35.55& 0.982& 6.14& 36.82& 0.983& 3.54& 32.46& 0.957& 4.28\\
& ShadowDiffusion &DHAN& 38.72& 0.987& 4.98& 37.78& 0.985& 3.44& 34.73& 0.970&3.63\\
& Inpaint4Shadow &DHAN& 36.73& 0.985& 5.70& 36.70& 0.985& 3.27& 33.27& 0.967&3.81\\
& DeS3 &N/A& 37.91& 0.986& 5.27& 37.45& 0.984& 3.03& 34.11& 0.968&3.56\\
& Homoformer  &DHAN& \underline{38.81}& \underline{0.987}& \textbf{4.25}& \textbf{39.45}& \underline{0.988}& \underline{2.85}& \underline{35.37}& \underline{0.972}& \underline{3.33}\\
& SAM-helps-shadow  &N/A& 33.94& 0.979& 7.44& 33.85& 0.981& 3.74& 30.72& 0.952& 4.79\\
& ours  &N/A& \textbf{39.08}& \textbf{0.989}& \underline{4.33}& \underline{39.36}& \textbf{0.992}& \textbf{2.58}& \textbf{35.57}& \textbf{0.975}& \textbf{3.11}\\
\hline
\multirow{8}{*}{\rotatebox[origin=c]{90}{ISTD+}} 
& BMNet&GT& 37.87&  0.991&  5.62& 37.51&  0.985&  2.45& 33.98&  0.972&  2.97\\
& ShadowDiffusion &GT& 39.69&  0.992& 4.97& 38.89&  0.987&  2.28& 35.67&  0.975&  2.72\\
& HomoFormer &GT& 39.49&  0.993& 4.73& 38.75&  0.984&  2.23& 35.35&  0.975&  2.64\\
\cline{2-12}
& \rule{0pt}{0.5ex}DC-ShadowNet &N/A& 31.06&  0.976& 12.62& 27.03&  0.961&  6.82& 25.03&  0.926&  7.77\\
& DeS3 &N/A& 36.49& 0.989& 6.56& 34.70& 0.972& 3.40& 31.38& 0.958& 3.94\\
& BMNet &FDRNet& -&-& 6.1& -& -&2.9& -& -& 3.5\\
& ShadowDiffusion &FDRNet & \underline{40.12}& \underline{0.992}& \underline{5.15}& \underline{36.66}& \underline{0.978}& \underline{2.74}& \underline{34.08}& \underline{0.968}& \underline{3.12}\\
& HomoFormer &FDRNet & 38.84& 0.991& 5.31& 34.58& 0.966& 3.17& 32.41& 0.953& 3.51\\
& Ours  &N/A& \textbf{40.36}& \textbf{0.993}& \textbf{4.78}& \textbf{37.89}& \textbf{0.982}& \textbf{2.46}& \textbf{35.00}& \textbf{0.972}& \textbf{2.85}\\
\hline\hline
\end{tabular}
}
\caption{The quantitative results of shadow removal using our SoftShadow and recent methods on SRD and ISTD+ datasets. The ``Input Masks" column shows the different types of input masks used by these methods. ``N/A" means the method does not require masks as input.
``GT" means the method uses manually annotated ground truth mask as input.
``DHAN" means using masks generated by DHAN~\cite{cun2020towards} method. ``FDRNet" means using masks generated by FDRNet~\cite{zhu2021mitigating} method. The best and the second results are \textbf{boldfaced} and \underline{underlined}, respectively.}
\label{Main Table}
\end{table*}

\begin{table}[t]
\centering
\setlength{\tabcolsep}{1mm}
\small
\resizebox{0.47\textwidth}{!}
{%
\begin{tabular}{c|ccc|ccc}
\hline\hline
\multirow{2}{*}{Methods} & \multicolumn{3}{c|}{LRSS} & \multicolumn{3}{c}{UIUC} \\
\cline{2-7}
 & PSNR$\uparrow$& SSIM$\uparrow$& MAE$\downarrow$& PSNR$\uparrow$& SSIM$\uparrow$& MAE$\downarrow$\\
\hline
DC-ShadowNet& 20.89& 0.902& 12.55& 24.85& 0.849&  9.51\\
G2R-ShadowNet& 20.90& 0.901& 9.99& 27.56& 0.858& 7.43\\
BCDiff& 22.13& 0.922& 10.68& 26.81&  0.852& 7.96\\
Ours& \textbf{23.32}& \textbf{0.933}& \textbf{9.77}& \textbf{28.85}& \textbf{0.903}& \textbf{6.48}\\
\hline\hline
\end{tabular}
}
\caption{The quantitative results of shadow removal using our SoftShadow and recent methods on the LRSS and UIUC datasets. 
The best results are \textbf{boldfaced}.}
\label{LRSS UIUC table}
\end{table}

\begin{table}[t!]
\centering
\setlength{\tabcolsep}{1mm}
\small 
\resizebox{0.25\textwidth}{!}
{
\begin{tabular}{c|cc}
\hline\hline
Methods& PSNR$\uparrow$ & MAE$\downarrow$ \\
\hline
Inpaint4Shadow & 40.10 & 4.23 \\
DeS3& 40.91& 4.08\\
HomoFormer& 40.82& 3.91\\
Ours & \textbf{41.84} & \textbf{3.77} \\
\hline\hline
\end{tabular}
}
\caption{The quantitative results in the penumbra area. The penumbra area is calculated from our predicted soft masks. The best results are \textbf{boldfaced}.}
\label{boundary}
\end{table}

\begin{table}[t!]
\centering
\setlength{\tabcolsep}{1mm}
\small
\resizebox{0.44\textwidth}{!}{
\begin{tabular}{c|cc|c}
\hline\hline
Input Masks& ShadowDiffusion & HomoFormer &Ours\\
\hline
Ground Truth mask & \underline{\textbf{35.67}} & \underline{35.35} & \multirow{5}{*}{35.00} \\
Otsu mask& \underline{35.62} &33.47 & \\
\cline{1-3}
BDRAR mask& 33.41& 32.96& \\
FDRNet mask& 34.08& 32.41 & \\
Pretrained SAM mask & 28.45 & 28.16 &  \\
\hline
Mean & 33.45& 32.47 &N/A\\
Std Dev & 0.912& 0.993&N/A\\
\hline\hline
\end{tabular}
}
\caption{The mask sensitivity evaluation on ISTD+ dataset. The values in the table are PSNR results. The best results are \textbf{boldfaced}. The results that are higher than ours are \underline{underlined}. }
\label{ISTD+}
\end{table}

\section{Experiments}

\subsection{Experimental Setups}

\paragraph{Implementation details} 
The shadow removal backbone in our SoftShadow framework can be any shadow removal network. We use ShadowDiffusion~\cite{guo2023shadowdiffusion} as the backbone example for all experiments.
We finetune ShadowDiffusion at a resolution of 256 $\times$ 256, followed by previous methods~\cite{guo2023shadowdiffusion}. We empirically set the threshold $t=0.76$ to get the ground truth soft shadow mask $\mathbf{s}_{gt}$. We employ the ViT-H model as the backbone for SAM~\cite{kirillov2023segment}. A set of LoRA~\cite{hu2021lora} layers with a rank of 8 is added to the self-attention blocks in the image encoder of SAM. We use Adam~\cite{kingma2014adam} optimizer. The training batch size is 16. We set $\lambda_1=0.1$ and $\lambda_2=1$. For evaluation, we use the DDIM sampler~\cite{song2020denoising} and 5 diffusion sampling steps. 
For more training details, please refer to the \textbf{supplementary}.

\vspace{1mm}
\noindent\textbf{Benchmark datasets} We work with four benchmark datasets for the various shadow removal experiments. \textbf{SRD Dataset}~\cite{qu2017deshadownet} consists of 2,680 training pairs and 408 testing pairs of shadow and shadow-free images. Notably, the SRD dataset does not provide shadow masks, the previous methods using SRD commonly using masks detected by DHAN \cite{cun2020towards} methods. 
\textbf{LRSS Dataset}~\cite{gryka2015learning} is a specifically designed soft shadow dataset that includes 137 images. We select 48 paired shadow and shadow-free images as our testing set. \textbf{UIUC Dataset} \cite{guo2012paired}   contains 76 pairs of images for testing. It features a variety of shadow types~\cite{jin2024des3}, including soft, hard, and self shadows, which provides a diverse and challenging set of conditions for evaluating shadow removal methods.
\textbf{ISTD+ Dataset}~\cite{le2019shadow} is adjusted ISTD dataset~\cite{wang2018stacked}, consists of 1330 training pairs and 540 testing pairs of shadow and shadow-free images. ISTD+ has manually annotated binary masks.

\vspace{1mm}
\noindent\textbf{Evaluation metrics} Following previous works~\cite{cun2020towards,guo2023shadowdiffusion,xiao2024homoformer}, we employ the Peak Signal-to-Noise Ratio (PSNR), Structural Similarity Index (SSIM) \cite{wang2004image}, and Mean Absolute Error (MAE)~\cite{willmott2005advantages} as quantitative evaluation metrics. We also calculate all metrics for shadow areas, non-shadow areas, and all pixels between ground truth shadow-free images and generated removal results. 


\subsection{Comparison with State-of-the-Art}
We compare our proposed method with several state-of-the-art shadow removal methods. We include methods that do not require shadow masks as input, e.g., DC-shadowNet~\cite{jin2021dc}, DeS3~\cite{jin2024des3}, and SAM-helps-shadow~\cite{zhang2023sam}. And methods that require shadow masks, including DHAN~\cite{cun2020towards}, BMNet~\cite{zhu2022bijective}, SG-ShadowNet~\cite{wan2022style}, ShadowFormer~\cite{guo2023shadowformer}, Inpaint4Shadow~\cite{li2023leveraging}, ShadowDiffusion~\cite{guo2023shadowdiffusion} and Homoformer \cite{xiao2024homoformer}. To demonstrate the generalizability of our method, we also compare it with zero-shot methods including BCDiff~\cite{guo2023boundary} and G2R-ShadowNet~\cite{liu2021shadow} on the LRSS~\cite{gryka2015learning} and UIUC~\cite{guo2012paired} datasets. 

\vspace{1mm}
\noindent\textbf{Quantitative results}
Table \ref{Main Table} shows the quantitative results of SRD and ISTD+ datasets. The SRD dataset contains many soft shadow images, which is more suitable for validating our methods. Specifically, we outperform all competing methods over all metrics, whether they require input shadow masks or not. 
Compared with the most recent work without requiring mask input, i.e., DeS3~\cite{jin2024des3}, the PSNR is improved from 34.11 dB to 35.57 dB in whole images. 
When compared with methods that need input masks, 
we outperform the SOTA method Homoformer~\cite{xiao2024homoformer}, which uses DHAN~\cite{cun2020towards} mask as their input. 
On the ISTD+ dataset, we significantly outperform DeS3~\cite{jin2024des3}, increasing the PSNR from 31.38 dB to 35.00 dB. 
For methods that rely on the ground truth masks from the ISTD+ dataset, we evaluate them using masks detected by the FDRNet~\cite{zhu2021mitigating} followed by previous methods BMNet~\cite{zhu2022bijective}, to ensure a fair comparison.
We outperform all competing methods under the condition of without ground truth manually annotated masks. 
Besides, we even achieve comparable results against some SOTA methods using ground truth masks.


Table \ref{LRSS UIUC table} shows the generalizability of our methods. We use our pretrained model on the SRD dataset and test it on LRSS and UIUC datasets without further training. The DeS3~\cite{jin2024des3} has better results because they train their method on the LRSS training set. 
For comparison, we chose DC-ShadowNet as our baseline since this method does not require masks as input. 
Our results significantly outperform DC-ShadowNet on LRSS and UIUC datasets. Additionally, we compare our methods with two methods, which are designed to have better generalizability. The results demonstrate that our method has the best PSNR/SSIM/MAE among the comparison methods.
These results highlight the robustness and adaptability of our model, even on datasets it was not specifically trained on.
\begin{figure*}[t!]
    \centering
    \includegraphics[width=0.89\textwidth]{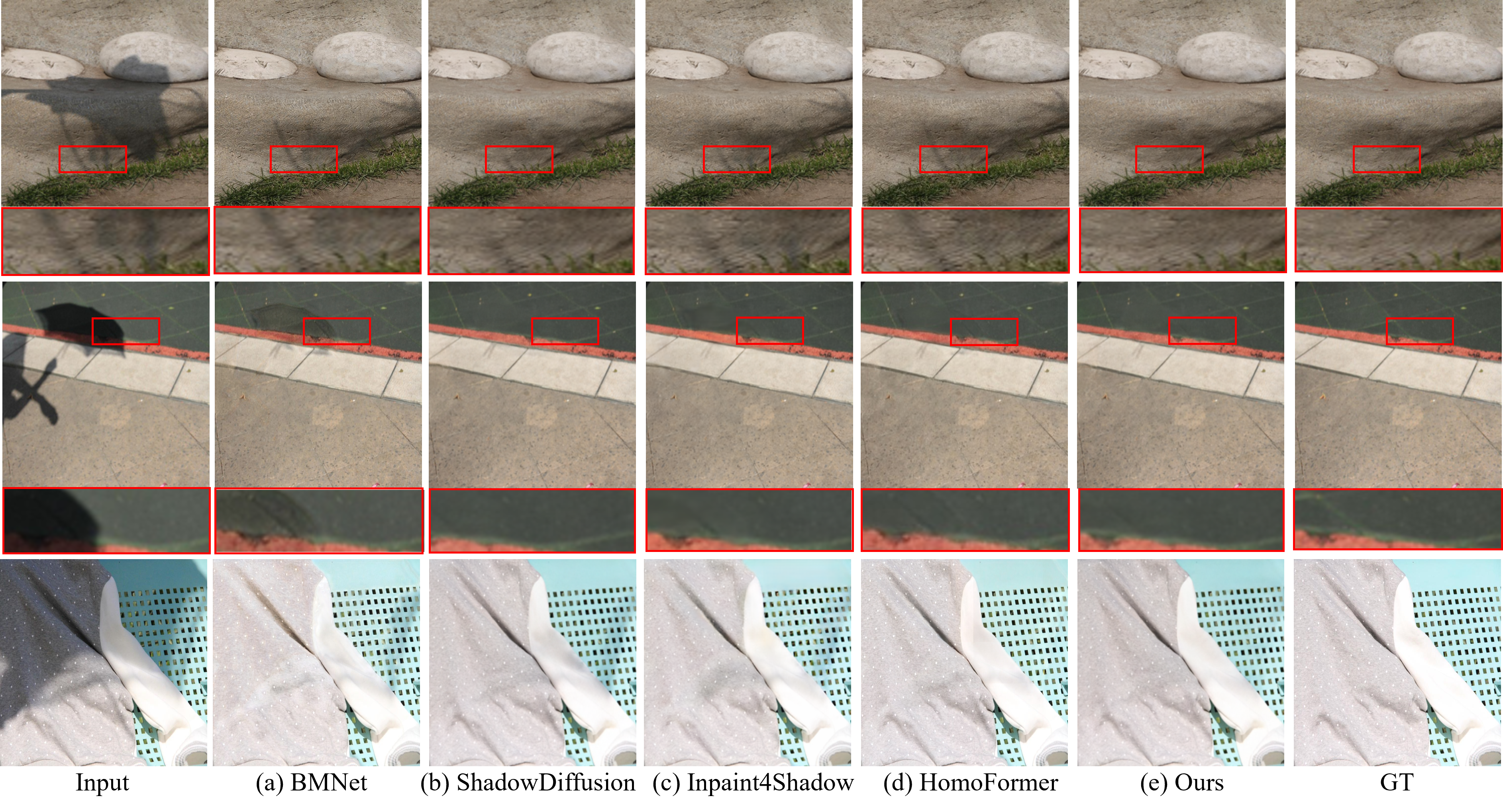}
    \caption{Examples of soft shadow image removal results on the SRD dataset~\cite{qu2017deshadownet}. The input shadow image, the estimated results of (a) BMNet~\cite{zhu2022bijective}, (b) ShadowDiffusion~\cite{guo2023shadowdiffusion}, (c) Inpaint4Shadow~\cite{li2023leveraging}, (d) Homoformer~\cite{xiao2024homoformer}, and (e) Ours, as well as the ground truth (GT) image, respectively.}
    \label{visual_SRD_1}
\end{figure*}

\begin{figure}[t!]
    \centering
    \includegraphics[width=0.43\textwidth]{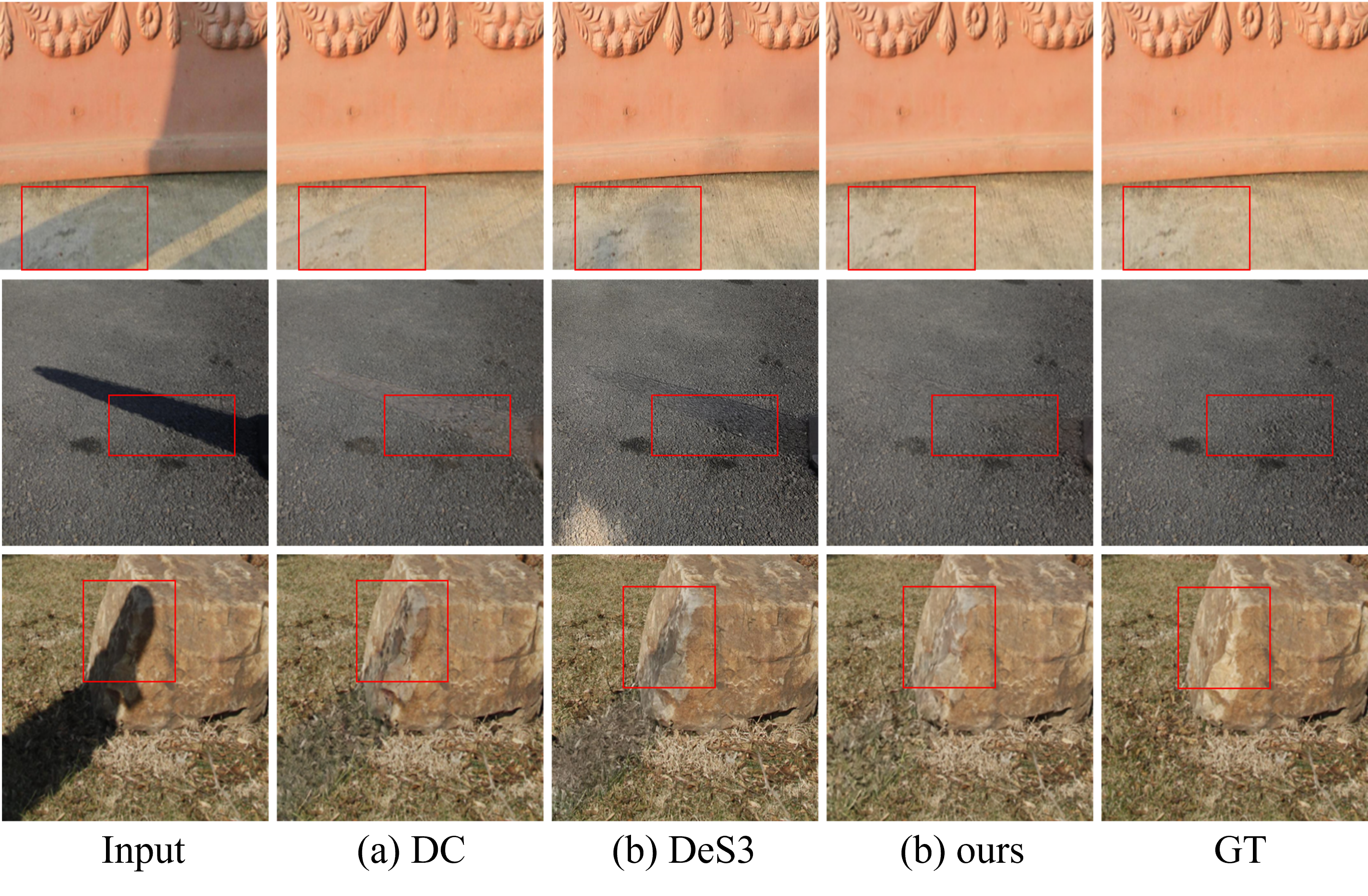}
    \caption{Examples of soft shadow removal results on SRD datasets~\cite{qu2017deshadownet}. The input shadow image, the estimated results of (a) DC-ShadowNet~\cite{jin2021dc}, (b) DeS3~\cite{jin2024des3}, and (c) Ours, as well as the ground truth image, respectively.}
    \label{visual_SRD_2}
\end{figure}

\begin{figure}[t!]
    \centering
    \includegraphics[width=0.47\textwidth]{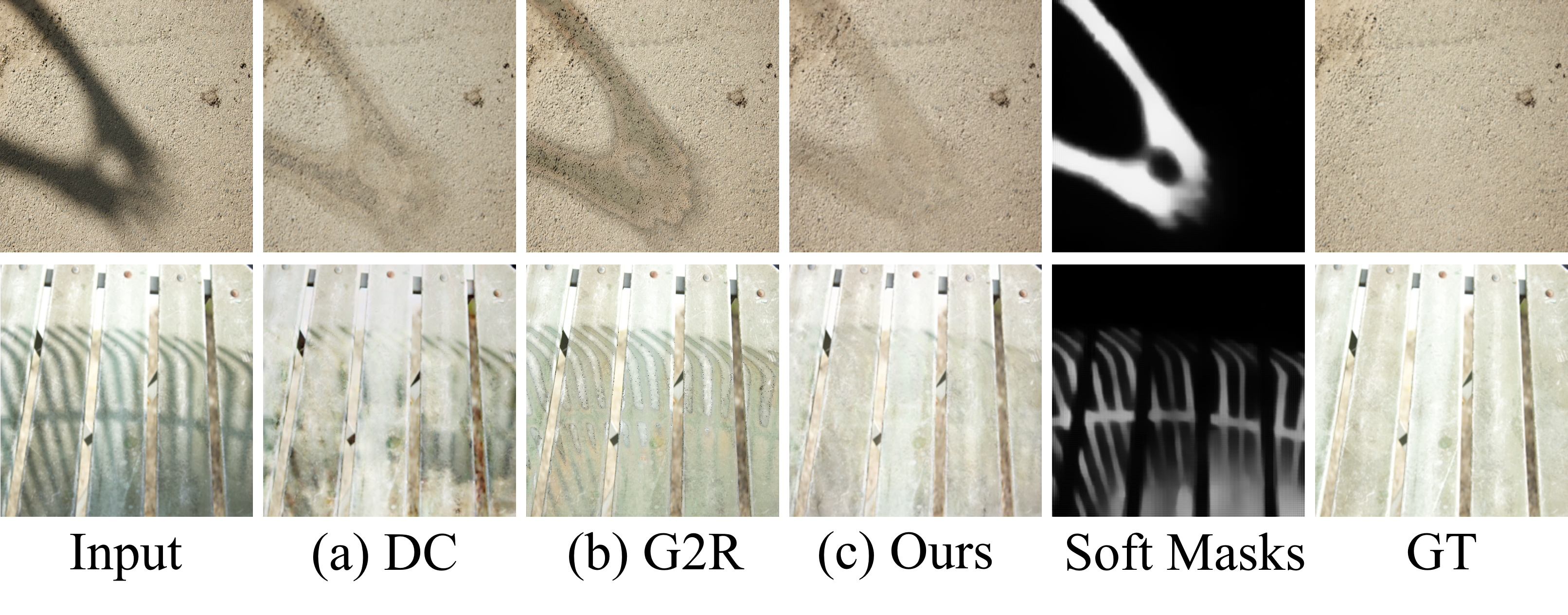}
    \caption{Examples of shadow removal results on the LRSS dataset~\cite{gryka2015learning}. The input shadow image, the estimated results of (a) DC-ShadowNet~\cite{jin2021dc}, (b) G2R-ShadowNet~\cite{liu2021shadow}, (c) Ours, our predicted soft shadow masks and ground truth image, respectively.}
    \label{LRSS}
\end{figure}

\vspace{1mm}
\noindent\textbf{Qualitative evaluation}
To further demonstrate the effectiveness of our methods compared to other competing methods, Figure~\ref{visual_SRD_1} illustrates our improvements in removing soft shadows on the SRD dataset compared with methods using hard shadow masks. The soft masks typically appear as soft boundaries of large object shadows or in small shadow regions. Specifically, our samples demonstrate that we effectively remove both types of soft shadows while preserving the illumination transition in penumbra areas, without introducing obvious artifacts. Figure~\ref{visual_SRD_2} shows our improvement in shadow area accuracy compared to other end-to-end methods. Our approach effectively removes all parts of the shadow shown in the first row of the example. In the second and third rows, we successfully remove shadows in regions with complex textures. Figure~\ref{LRSS} highlights our method's generalizability on the LRSS dataset, achieving improved shadow removal performance without additional training on LRSS dataset. These results confirm our method's adaptability and precision in diverse shadowed scenarios.

\section{Framework Analysis}
\label{mask_sensitivity}
In this section, we provide more experiments to evaluate the effectiveness of our proposed SoftShadow framework. We discuss the performance on the penumbra area, the mask sensitive of previous methods, and we discuss the effectiveness of our provided losses in ablation study.

\noindent\textbf{Penumbra effect evaluation}
The penumbra region typically occupies only a small portion of the entire image, so improvements within this area have a limited impact on overall image metrics. To further assess the improvements introduced by using soft masks in the penumbra region, we calculate the PSNR and MAE values specifically within this area on the SRD dataset. For a fair comparison, we define the penumbra region following the methods from AEF~\cite{fu2021auto}, applying soft masks generated by our method. The results are shown in Table~\ref{boundary}.
Compared with the SOTA methods like HomoFormer~\cite{xiao2024homoformer}, DeS3~\cite{jin2024des3} and Inpaint4Shadow~\cite{li2023leveraging}, we have improvements on PSNR for around 1db. These results show that on images containing soft shadows, our methods has better restoration performances.

\vspace{1mm}
\noindent\textbf{Mask sensitivity evaluation}
Previous shadow removal methods that rely on external shadow masks are sensitive to the accuracy of the input masks. Their performance can degrade when lack of ground truth mask. In contrast, end-to-end shadow removal methods mitigate the impact of low-quality masks. To further illustrate the sensitivity of prior SOTA methods to input masks, we evaluate their performance using various mask inputs. 

Specifically, we evaluate ShadowDiffusion~\cite{guo2023shadowdiffusion} and HomoFormer~\cite{xiao2024homoformer} using a range of input masks, including ground truth (GT) masks, Otsu masks~\cite{hu2019mask}, BDRAR~\cite{zhu2018bidirectional} detected masks, FDRNet~\cite{zhu2021mitigating} detected masks and pretrained SAM~\cite{kirillov2023segment} detected masks as inputs.  The PSNR results on the ISTD+~\cite{le2019shadow} datasets are shown in Table~\ref{ISTD+}. The Otsu mask follows the approach in Mask-ShadowGAN~\cite{hu2019mask}, using the difference between shadow-free and shadowed images to generate masks; both Otsu and GT masks include additional shadow-free information. Results indicate that our methods maintain consistent performance across all tests, whereas ShadowDiffusion and HomoFormer have performance declines when using generated masks. Although our method achieves slightly lower PSNR values than those obtained with Otsu or GT masks, this is reasonable given that the ISTD+ dataset contains a larger proportion of hard shadow images, where our method has less of an advantage. Our approach surpasses the rest results that rely on masks generated by shadow detectors. 



\vspace{1mm}
\noindent\textbf{Ablation Study}
We ablate our method, SoftShadow, by adding three losses:  the shadow removal loss $\mathcal{L}_{rem}$, the mask reconstruction loss $\mathcal{L}_{mask}$,  and the penumbra formation constraint loss $\mathcal{L}_{pen}$, and demonstrate the results on SRD and LRSS datasets. First, we present the shadow removal results using the pretrained SAM and ShadowDiffusion models. Next, we finetune the framework with the shadow removal loss $ \mathcal{L}_{rem}$. Then, we add the mask reconstruction loss $\mathcal{L}_{mask}$ to provide SAM with more precise mask position guidance. Finally, we add the penumbra formation constraint loss $\mathcal{L}_{pen}$ to refine the mask boundaries for smoother transitions in the shadow regions. 

\begin{figure}[t!]
    \centering
    \includegraphics[width=.47\textwidth]{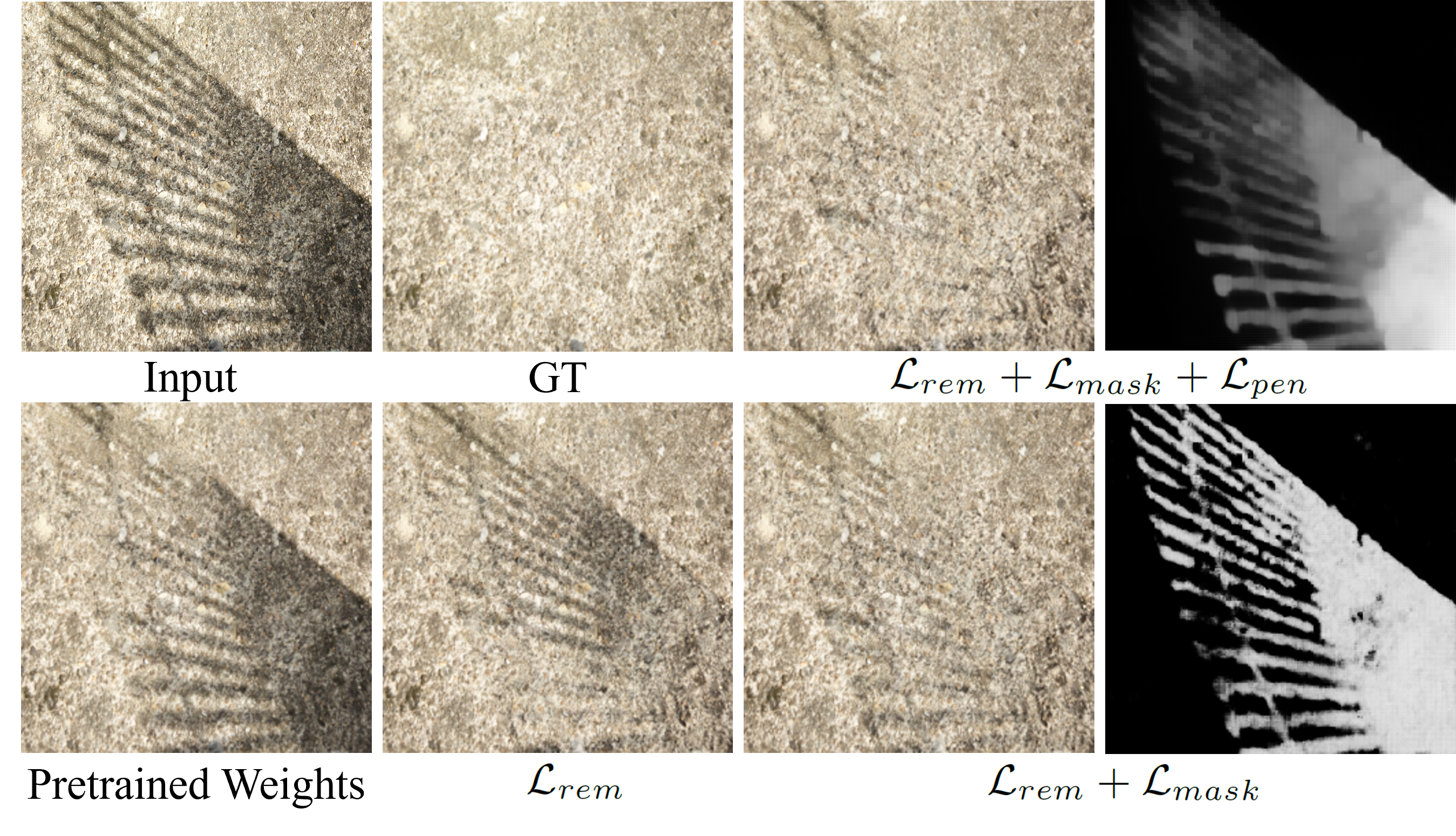}
    \caption{Visual comparison of results from our framework with different configurations: using pretrained weights, finetuned with the loss function $\mathcal{L}_{rem}$, $\mathcal{L}_{rem} + \mathcal{L}_{mask}$ and our full configuration $\mathcal{L}_{rem} + \mathcal{L}_{mask} + \mathcal{L}_{pen}$ (Ours), respectively. }
    \label{ablation}
\end{figure}

The results in Table~\ref{ablation table} are based on training with the SRD training set and evaluating on both the SRD and LRSS datasets. 
As shown in Table~\ref{ablation table}, the shadow removal loss $\mathcal{L}_{rem}$ improves performance on the SRD dataset, primarily by enhancing the shadow removal network. Adding the mask reconstruction loss $\mathcal{L}_{mask}$ further fine-tunes SAM, producing more accurate shadow masks. This improvement of mask accuracy leads to better shadow removal results. To evaluate the impact of the penumbra formation constraint loss $ \mathcal{L}_{pen}$, we compare results with and without $ \mathcal{L}_{pen}$. The results show that $ \mathcal{L}_{pen}$ further aids in shadow removal on the SRD dataset, refining the shadow edges and enhancing the overall shadow removal quality. 
In the LRSS dataset, both $\mathcal{L}_{rem}$  and $ \mathcal{L}_{mask}$ significantly improve performance, the $ \mathcal{L}_{pen}$ improves the PSNR and MAE significantly, although it results in a slight decrease in SSIM. As shown in Figure~\ref{ablation}, adding the constraint loss $ \mathcal{L}_{pen}$ leads to more accurate and smoother soft mask predictions. The combined effects of $ \mathcal{L}_{mask}$ and $ \mathcal{L}_{pen}$ lead to more accurate and smoother soft mask predictions. Together, these losses provide the removal framework with enhanced soft shadow details, resulting in superior performance in shadow removal.

\begin{table}[t]
\centering
\setlength{\tabcolsep}{1 mm}
\small
\resizebox{0.47\textwidth}{!}{
\begin{tabular}{c|ccc|ccc}
\hline\hline
\multirow{2}{*}{Methods} & \multicolumn{3}{c|}{LRSS} & \multicolumn{3}{c}{SRD} \\
\cline{2-7}
 & PSNR$\uparrow$& SSIM$\uparrow$& MAE$\downarrow$& PSNR$\uparrow$& SSIM$\uparrow$& MAE$\downarrow$\\
\hline
Pretrained Weights& 21.40& 0.910& 12.26& 31.27& 0.963& 4.41\\
$\mathcal{L}_{rem}$ & 21.78& 0.913& 12.07& 35.27& 0.973& 3.32\\
$\mathcal{L}_{rem} + \mathcal{L}_{mask}$& 23.08& \textbf{0.935}& 9.97& 35.44& 0.974& 3.17\\
$\mathcal{L}_{rem} + \mathcal{L}_{mask} + \mathcal{L}_{pen}$& \textbf{23.32}& 0.933& \textbf{9.77}&\textbf{ 35.57}& \textbf{0.975}& \textbf{3.11}\\
\hline\hline
\end{tabular}
}
\caption{The ablation studies for the shadow removal loss $\mathcal{L}_{rem}$, the mask reconstruction loss $\mathcal{L}_{mask}$, and the penumbra formation constraint loss $\mathcal{L}_{pen}$ in LRSS and SRD datasets. The best results are \textbf{boldfaced}.}
\label{ablation table}
\end{table}


\section{Conclusion}
In this paper, we introduce novel soft shadow masks designed specifically for shadow removal. To achieve this, we propose a unified framework, \textit{SoftShadow}, that eliminates the need for additional shadow mask input. By leveraging a pretrained SAM with LoRA, the framework accurately predicts soft masks as intermediate results, capturing detailed and varied shadow information. Additionally, we introduce a penumbra formation constraint, inspired by the physical shadow formation model, to jointly tune SAM and the shadow removal network, refining the soft mask in the penumbra area and facilitating artifact-free restoration.
Extensive experiments demonstrate that our method is superior on various occasions, proving the validity of our method.

\clearpage
{
    \small
    \bibliographystyle{ieeenat_fullname}
    \bibliography{main}
}



    
\end{document}